\documentclass[11pt]{article}

\usepackage[utf8]{inputenc}
\usepackage[T1]{fontenc}
\usepackage{xcolor}
\usepackage{graphicx}
\usepackage{subfig}
\usepackage{wrapfig}
\usepackage[hidelinks]{hyperref}
\usepackage{url}
\usepackage{booktabs} 
\usepackage{abstract}
\usepackage{multicol}
\usepackage{fancyhdr}
\usepackage{amsmath}
\usepackage{amsfonts}
\usepackage{amsthm}
\usepackage{amssymb}
\usepackage{nicefrac}
\usepackage{microtype}
\usepackage{paralist}
\usepackage[font=small,labelfont=bf]{caption}
\usepackage[ruled,norelsize]{algorithm2e}
\usepackage[noend]{algorithmic}
\usepackage{calc}
\usepackage{bbm}
\usepackage{enumitem}
\usepackage{scalerel}

\allowdisplaybreaks

\newcommand{\R}{\mathbb{R}}

\newcommand{\Normal}{\mathcal{N}}

\DeclareMathOperator{\spn}{span}

\newcommand{\bmat}{\begin{pmatrix}}
\newcommand{\emat}{\end{pmatrix}}

\newcommand{\mstr}[1]{\mathrm{#1}}

\begin{document}

\title{Variable Metric Evolution Strategies\\for High-dimensional Multi-Objective Optimization}

\pagestyle{fancy}
\fancyhead{}
\fancyfoot{}
\fancyhead[L]{Variable Metric ESs for High-dim.\ Multi-Objective Optimization}
\fancyhead[R]{T.~Glasmachers}
\fancyfoot[C]{\thepage}
\fancypagestyle{firststyle}
{
   \fancyhf{}
   \fancyfoot{}
}

\author{
	Tobias Glasmachers\\
	Institute for Neural Computation\\
	Ruhr-University Bochum, Germany\\
	\texttt{tobias.glasmachers@ini.rub.de}
}

\date{}

\maketitle

\begin{abstract}
We design a class of variable metric evolution strategies well suited
for high-dimensional problems. We target problems with many variables,
not (necessarily) with many objectives. The construction combines two
independent developments: efficient algorithms for scaling covariance
matrix adaptation to high dimensions, and evolution strategies for
multi-objective optimization. In order to design a specific instance of
the class we first develop a (1+1) version of the limited memory matrix
adaptation evolution strategy and then use an established standard
construction to turn a population thereof into a state-of-the-art
multi-objective optimizer with indicator-based selection. The method
compares favorably to adaptation of the full covariance matrix.
\end{abstract}

\section{Introduction}

Multi-objective optimization deals with the simultaneous optimization
(w.l.o.g.\ minimization) of multiple conflicting objective functions
$f_1, \dots, f_m : X \to \R$. A common optimization goal suitable for
decision support is to approximate the Pareto optimal set, which
consists of all non-dominated solutions. A solution $x$ (weakly)
dominates $x'$ if $f_i(x) \leq f_i(x')$ for all $i \in \{1, \dots, m\}$.
We can think of the Pareto set as the set of optimal compromises, while
its image under $f$ is called Pareto front. This set shall be
approximated in a three-fold manner: solutions are expected be as close
as possible to the set, they shall ``cover'' the set well in the sense
of being representative of qualitatively different optimal trade-offs,
and the solution set is often requested to be of a pre-specified maximal
cardinality. If we are restricted to black-box access to the objective
functions then such problems are commonly solved with optimization
heuristics, in particular with multi-objective evolutionary algorithms
(MOEAs).

In this paper we focus on the case that the search space $X$ is
(embedded in) a relatively high-dimensional continuous search space
$\R^n$. Dealing with hundreds or even thousands of decision variables
($n \gg 100$) can pose a significant challenge to black-box direct
search algorithms. The setup is demanding even for single-objective
optimization for the following reasons. First of all, runtime (measured
in the number of objective function evaluations, hereafter referred to
as sample complexity or algorithm external cost) of black-box direct
search methods scales with $n$, usually at least linearly
\cite{teytaud2006general,jagerskupper2006quadratic,akimoto2018drift}. In
order to obtain a close-to-optimal convergence rate, efficient
state-of-the-art search strategies like CMA-ES
\cite{hansen2001completely} adapt the covariance matrix of the search
distribution to the problem at hand. The matrix consists of
$\Theta(n^2)$ scalar parameters, resulting in an undesirable algorithm
internal cost per sample of order $\Omega(n^2)$. Furthermore, the sample
complexity of learning the covariance information scales super-linear.
This results in a runtime of $\omega(n^2)$ of the transient phase in
which the covariance matrix is learned. Many alternative methods pay a
similar (yet implicit and less well analyzed) price through the need to
work with very large populations.

For single-objective optimization that problem was addressed by
developing special purpose algorithms from the class of evolution
strategies
\cite{sun2013linear,loshchilov2014computationally,akimoto2016projection,loshchilov2018large}.
Their common theme is that they abandon the full covariance matrix in
favor of more scalable but restricted representations of multi-variate
normal distributions. While that design reduces the algorithm internal
costs per sample, it can incur additional algorithm external costs if
the restricted covariance matrix does not allow to approximate the
optimal covariance matrix sufficiently well. For large $n$, the
resulting trade-off between algorithm internal and external complexity
is often favorable.

\paragraph{Contributions}

In the present paper we aim to make the above sketched development
available for multi-objective optimization. Our proceeding and the
corresponding contributions are as follows:
\begin{itemize}
\item
	We design a general blueprint of an MOEA for high-dimensional
	search spaces with indicator-based selection.
\item
	We design a minimal elitist variant of the limited memory matrix
	adaptation evolution strategy (LM-MA-ES) \cite{loshchilov2018large},
	called (1+1)-LM-MA-ES.
\item
	We instantiate the algorithm template using (1+1)-LM-MA-ES,
	resulting in the novel MO-LM-MA-ES algorithm.
\item
	In the spirit of open and reproducible research we publish the full
	source code of our implementation as a supplement to this paper.
\item
	For the purpose of evaluating our algorithm we make a Python version
	of the benchmark collection presented in
	\cite{glasmachers2019challenges} available.
\end{itemize}

\paragraph{Existing Methods}

A recent overview of multi-objective EAs for high-dimensional problems
is found in \cite{tian2021evolutionary}. The review article
distinguishes various classes of methods of which we briefly discuss the
most prominent ones. For details we refer to the extensive collection of
references therein.

Many existing techniques approach the problem by performing a
dimensionality reduction of some sort. Such methods can work well
(theoretically and practically) if and only if the problem has an
inherent low-dimensional structure. We argue that that is a very strong
assumption on a black-box problem. If violated, and in particular if the
subspace is poorly chosen, then the algorithm may not even have a chance
to approach the Pareto front, irrespective of its compute budget.

Another popular technique is grouping of variables into independent
blocks. If such a problem decomposition is known then it should of
course be exploited, essentially giving rise to independent sub-problems
of tractable dimensionality. Designing such a decomposition obviously
requires white-box access to the problem. In a black-box scenario one
may still learn the structure on the fly. However, the very assumption
of the existence of such a structure is hard to justify in general. Even
if a decomposition exists it is unclear why it should involve groups of
variables, in contrast to general subspaces. Performance can be expected
to drop when the assumption is violated.

Both of the above described approaches make strong assumptions about the
black-box problem at hand. The assumptions are chosen to match the
proposed methods, not the experience with actual application problems.
We argue that that is a dangerous approach when addressing black-box
optimization, and that there is a need for generic yet efficient
multi-objective solvers. Such an approach should leverage existing work
on scaling up single-objective algorithms to high-dimensional search
spaces.

\section{Background}

In this section we introduce the background that is necessary for
following the construction of our new algorithms. We briefly review two
existing algorithms, namely the multi-objective covariance matrix
adaptation evolution strategy (MO-CMA-ES) and the limited memory matrix
adaptation evolution strategy (LM-MA-ES). Both methods will serve as
blueprints for the new algorithms presented in the next section.

\subsection{Multi-Objective CMA-ES}

The multi-objective covariance matrix adaptation evolution strategy
(MO-CMA-ES) \cite{igel2007covariance} is a state-of-the-art MOEA for
continuous optimization. It is constructed as a rather generic MOEA with
indicator-based selection \cite{zitzler2004indicator}. As such, several
variants exist. Selection is a two-stage process based on dominance, or
algorithmically speaking, on non-dominated sorting. A secondary
selection criterion is applied within the non-dominance ranks. While the
crowding distance can be used in principle, modern implementations
usually rely on hypervolume contributions \cite{igel2008shark}. A
further improvement over the base version is a refined notion of
success for implementing the $1/5$ step size adaptation rule
\cite{voss2010improved}. MO-CMA-ES usually employs elitist schemes,
either using $(\mu+\mu)$ or $(\mu+1)$ selection. The latter is also
known as steady-state selection.

\begin{algorithm*}
\caption{MO-CMA-ES}
\label{MO-CMA-ES}
\begin{algorithmic}[1]
\STATE{\textbf{initialize} (1+1)-CMA-ES instances $x_i$ for $i \in \{1, \dots, \mu\}$}
\REPEAT
	\FOR{$i \leftarrow 1, \ldots, \mu$}
		\STATE{select parent $x_i$ with $i \in \{1, \dots, \mu\}$}
		\STATE{sample offspring $y_i$ from multi-variate normal distribution}
	\ENDFOR
	\STATE{rank population formed by parents $x_i$ and offspring $y_i$ by non-dominated\\sorting and hypervolume contributions}
	\FOR{$i \leftarrow 1, \ldots, \mu$}
		\IF{offspring $y_i$ is among the $\mu$ best individuals}
			\STATE{update the covariance matrix of parent and offspring}
		\ENDIF
		\STATE{update step size of parent and offspring based on success}
	\ENDFOR
	\STATE{keep the best $\mu$ individuals, store them in $x_i$, $i \in \{1, \dots, \mu\}$}
\UNTIL{stopping criterion is met}
\end{algorithmic}
\end{algorithm*}

The MO-CMA-ES is based on a simplified elitist CMA-ES variant designed
specifically for that purpose, the (1+1)-CMA-ES. It maintains a
population of $\mu$ independent instances. In other words, each
individual in MO-CMA-ES can be though of as a fully fledged (1+1)-CMA-ES
instance including strategy parameters. Depending on the population
scheme offspring are generated by either creating one offspring per
parent or by selecting a single parent at random. The environmental
selection step then reduces the overall population back to the previous
size of $\mu$ elitist instances. The approach is formalized in
algorithm~\ref{MO-CMA-ES}. The main purpose of the pseudo-code is to
illustrate the working principle of solving a multi-objective problem
with a population of instances of (1+1) algorithms. It therefore glosses
over several details that are non-essential for this discussion, like
the full state of the elitist optimizers (which we think of as being
attached to the search points) and the details of the update mechanisms
of step size and covariance matrix. We refer the interested reader to
\cite{igel2007covariance,voss2010improved} for details, and to
\cite{igel2008shark} for a reference implementation.

\subsection{Evolution Strategies for High-dimensional Problems}

During the last decade several evolution strategies (ESs) for
high-dimensional problems were developed. Before that only simple ESs
with isotropic search distributions as well as ESs with diagonal
covariance matrix were available for that case, since they enable
sampling of offspring and covariance matrix updates in linear time.
Such algorithms perform well only on separable problems, which can be
argued to be somewhat trivial since each variable can be optimized
independently. In contrast, fully fledged CMA-ES can handle arbitrary
dependencies between variables, but has algorithm internal costs of
$\Omega(n^2)$ per sample.

The newer development therefore aimed to solve non-separable problems
efficiently while keeping the algorithm internal costs tractable. The
route to success is to model only a small subset of all search
directions, and actually the most important ones. The can be achieved
by means of low-rank representations of square matrices, which are at
the heart of all existing methods
\cite{sun2013linear,loshchilov2014computationally,akimoto2016projection,loshchilov2018large}.
They effectively maintain a covariance matrix of the form
\begin{align}
	\label{eq:low-rank}
	C = I + \sum_{i=1}^k m_i m_i^T.
\end{align}
While sampling isotropically in $n - k$ directions, the algorithms
explicitly model the $k$ most relevant search directions. The rank $k$
of the adaptive part is usually chosen as
$k \in \Theta\big(\log(n)\big)$, yielding an overall complexity of
$\Theta\big(n \log(n)\big)$ per sample. An additional benefit is that
the reduced number of parameters allows for a larger learning rate and
hence faster adaptation. The obvious downside is reduced efficiency on
problems requiring a full covariance matrix for optimal performance.

A consequence of the construction is that the algorithms cannot be fully
invariant to affine transformations of the search space. They are,
however, invariant to translation, scaling and rotation, and (due to
ranking) to strictly monotone fitness transformations. Furthermore, most
adaptation rules are affine invariant to transformations \emph{within}
the represented subspace $\spn(\{m_1, \dots, m_k\})$. In this sense,
they are well aligned with the design principles underlying CMA-ES.

The VD-CMA-ES \cite{akimoto2016projection} combines a low-rank
representation with a diagonal matrix. The design has the advantage to
enable fast learning of independent scaling factors, at the expense of a
less elegant design and giving up some invariance properties.

On many high-dimensional problems the low-rank approach yields a
favorable compromise between algorithm internal cost (computational cost
per sample) and sample efficiency (number of function evaluations) as
compared with maintaining and adapting a full covariance matrix. It was
demonstrated in \cite{loshchilov2018large} that algorithm internal costs
(which can become dominant for large $n$) can be reduced by a factor of
roughly $10$ when working with hundreds to thousands of variables.

\begin{algorithm*}
\caption{LM-MA-ES}
\label{LM-MA-ES}
\begin{algorithmic}[1]
\STATE{\textbf{parameters}
		$\lambda = 4 + \lfloor 3 \log(n) \rfloor$;
		$\mu = \lfloor \lambda/2 \rfloor$;
		$w_i = \frac{\log(\mu + \frac{1}{2}) - \log(i)}{\sum^{\mu}_{j=1}(\log(\mu + \frac{1}{2})-\log(j))} \;$
		\\
		$\mstr{for} \; i=1, \ldots, \mu$;
		$\mu_w = \frac{1}{\sum^{\mu}_{i=1} w^2_i}$;
		$k = 4 + \lfloor 3 \log(n) \rfloor$;
		$c_{\sigma} = \frac{2 \lambda}{n}$;
		$c_{d,i} = \frac{1}{1.5^{i-1} n}$;
		\\
		$c_{c,i} = \frac{\lambda}{ 4^{i-1} n} \; \mstr{for} \; i=1, \ldots, k$
	}
\STATE{\textbf{initialize state} $x \in \R^{n},\enspace \sigma > 0,\enspace p_{\sigma} = 0$, $m_i \in \R^{n}$, $m_i = 0 \; \mstr{for} \; i=1, \ldots, k$}
\REPEAT
	\FOR{$i \leftarrow 1,\ldots,\lambda$}
		\STATE{$z_i \leftarrow {{\mathcal N}  \hspace{-0.13em}\left(0, I\,\right)}$}
		\STATE{$d_i \leftarrow z_i$}
		\STATE{\textbf{for} $j \leftarrow	 1,\ldots,k \; \textbf{do}$}
		\STATE{$\;\;\;\; d_i \leftarrow (1 - c_{d,j}) d_i + c_{d,j} m_{j}\left((m_{j})^T d_i  \right)$}
		\STATE{ $f_i \leftarrow f(x + \sigma d_i)$}
	\ENDFOR
	\STATE{$ x \leftarrow  x + \sigma \sum_{i=1}^{\mu} w_i d_{i:\lambda} \;$} \qquad\qquad\qquad \COMMENT{$i:\lambda$ denotes $i$-th best sample on $f$}
	\STATE{$ p_{\sigma} \leftarrow (1 - c_{\sigma}) p_{\sigma} + \sqrt{\mu_w c_{\sigma}(2-c_{\sigma})} \sum_{i=1}^{\mu} w_i z_{i:\lambda} $}
	\STATE{\textbf{for} $i \leftarrow 1,\ldots,k$ \textbf{do}}
	\STATE{$\;\;\;\; m_i \leftarrow (1-c_{c,i})m_i + \sqrt{\mu_w c_{c,i}(2-c_{c,i})}  \sum_{j=1}^{\mu} w_j z_{j:\lambda}$}
	\STATE{$\sigma \leftarrow \sigma \cdot \exp
	          \left[ \frac{c_{\sigma}}{2} \left(  \frac{\left\| p_{\sigma} \right\|^2}{ n} - 1  \right) \right] $} \label{MAStepSizeUpdate} 
\UNTIL{stopping criterion is met}
\end{algorithmic}
\end{algorithm*}

The limited memory matrix adaptation evolution strategy (LM-MA-ES) by
Loshchilov et al.\ \cite{loshchilov2018large} is one such method.
Algorithm~\ref{LM-MA-ES} shows the full pseudo code of LM-MA-ES. The
algorithm incorporates standard components of modern evolution
strategies like sampling from a multi-variate Gaussian (line~5),
rank-based weighted intermediate recombination (line~10), and cumulative
step size adaptation (lines 11 and 14). Its core mechanisms are the
construction of samples $d_i$ from isotropic Gaussian samples $z_i$
(lines 6 to 8) and the update of the direction vectors $m_i$ (line~13).
The direction vectors correspond to the covariance model of
equation~\eqref{eq:low-rank}. They are updated as evolution paths
with an exponentially decaying spectrum of learning rates, operating at
different time scales.

\section{Algorithms}

We now present two novel algorithms, namely a simplified elitist variant
of LM-MA-ES and, our core contribution, a multi-objective evolution
strategy for high-dimensional spaces.

\subsection{An Elitist LM-MA-ES}

The first step is to develop an elitist variant of LM-MA-ES. Its pseudo
code is found in algorithm~\ref{Elitist-LM-MA-ES}. The construction
follows the procedure of turning CMA-ES into an elitist algorithm known
as the (1+1)-CMA-ES \cite{igel2007covariance}. Here, it is applied to
LM-MA-ES with its efficient low-rank approach. Consequently, the new
algorithm closely follows the operation of LM-MA-ES, but with the
following changes:

\begin{itemize}
\item
	The algorithm employs (1+1) selection. Therefore, only a single
	offspring is sampled in each generation (line~4).
\item
	Step size control is success-based, similar to the original
	evolution strategy proposed by Rechenberg \cite{rechenberg1973}.
	We follow a common modern implementation~\cite{kern2004learning}
	(lines 10 and~12).
\item
	Covariance matrix updates are performed only in case of a successful
	offspring (line~9).
\item
	While maintaining the rank of order $k \in \Theta\big(\log(n)\big)$,
	the method requires different parameter settings to compensate for
	the availability of only a single offspring sample per generation
	(line~1).
\item
	We slightly change and actually simplify the encoding of the
	covariance matrix so as to enable parallelization of the offspring
	sampling procedure (line~5).
\end{itemize}

\begin{algorithm*}
\caption{(1+1)-LM-MA-ES}
\label{Elitist-LM-MA-ES}
\begin{algorithmic}[1]
\STATE{\textbf{parameters}
		$k = 4 + \lfloor 3 \log(n) \rfloor$;
		$\alpha^+ = n/2$; $\alpha^- = -n/8$;
		$c_{d,i} = \frac{1}{1.5^{i-1} n}$;
		\\
		$c_{c,i} = \frac{k}{ 4^{i-1} n} \; \mstr{for} \; i=1, \ldots, k$
	}
\STATE{\textbf{initialize state} $x \in \R^{n},\enspace \sigma > 0,\enspace m_i = 0 \in \R^{n} \mstr{ for } i=1, \ldots, k$}
\REPEAT
	\STATE{$z \leftarrow \Normal(0, I)$}
	\STATE{$d \leftarrow z + \sum_{i=1}^k (m_i^T z) m_i$}
	\STATE{$y \leftarrow x + \sigma d$}
	\IF{$f(y) \leq f(x)$}
		\STATE{$x \leftarrow y$}
		\STATE{\textbf{for} $i \leftarrow 1, \ldots, k \; \textbf{do} \; m_i \leftarrow (1-c_{c,i}) \cdot m_i + \sqrt{c_{c,i}(2-c_{c,i})} \cdot z$}
		\STATE{$\sigma \leftarrow \sigma \cdot \exp(\alpha^+)$}
	\ELSE
		\STATE{$\sigma \leftarrow \sigma \cdot \exp\left(\alpha^-\right)$}
	\ENDIF
\UNTIL{stopping criterion is met}
\end{algorithmic}
\end{algorithm*}

Compared with the population-based LM-MA-ES, the calculation of the
summands $(m_i^T z) m_i$ does not have any sequential dependency.
Therefore, line~5 can be implemented as $M z M^T$ (two vector-matrix
products), where the rows of the $k \times n$ matrix $M$ hold the path
vectors~$m_i$. A similar consideration applies to the vectorized
implementation of the evolution path update in line~9.

\subsection{Multi-objective LM-MA-ES}

With these ingredients in place the construction of a multi-objective
evolution strategy for high-dimensional problems becomes
straightforward. We exploit the fact that the pseudo-code of MO-CMA-ES
(see algorithm~\ref{MO-CMA-ES}) is nothing but a rather generic MOEA
with indicator based selection. For the new algorithm, we swap the
(1+1)-LM-MA-ES in for (1+1)-CMA-ES. We call the resulting algorithm the
multi-objective LM-MA-ES (MO-LM-MA-ES). We abstain from showing its
pseudo code since it coincides with algorithm~\ref{MO-CMA-ES} up to the
fact (1+1)-LM-MA-ES instances are used throughout as individuals.

It is well known that multi-objective optimization with significantly
more than two or three objectives (known as many-objective optimization)
can profit from a different algorithm design, as exercised by MOEA/D and
NSGA-III \cite{zhang2007moead,deb2013evolutionary}.
We assume that these algorithms can be adapted to leverage
(1+1)-LM-MA-ES for solving high-dimensional problems. Such an
investigation is outside of the scope of this paper.

\section{Experiments}

We performed an empirical study with the goal to assess the performance
of our new algorithms. More precisely we aim to answer the following
questions:
\begin{enumerate}
\item
	How does (1+1)-LM-MA-ES compare to its population-based counterpart?
\item
	How does MO-LM-MA-ES perform on high-dimensional problems?
\item
	How well does MO-LM-MA-ES approximate the Pareto front, and how does
	the approximation quality compare with using a full-rank covariance
	matrix?
\end{enumerate}
One could of course ask further questions. One such point is to compare
(1+1)-LM-MA-ES to less scalable full-rank variable metric methods.
However, the focus of this paper is multi-objective optimization. For
single-objective optimization, the relation between low-rank and
full-rank algorithms is already well studied in the literature, so we
refer to the corresponding studies
\cite{sun2013linear,loshchilov2014computationally,akimoto2016projection,loshchilov2018large}.

We make Python implementations of our algorithms available as an online
supplement to this paper. Furthermore, the code supplement contains our
Python port of the benchmark collection \cite{glasmachers2019challenges}
used in the multi-objective experiments. The code is available on the
author's website%
\footnote{\url{https://www.ini.rub.de/the_institute/people/tobias-glasmachers/\#software}
}.

We scale all objective functions so that the optimal values of each
objective is zero, and that in the single-objective optima the other
objective function has a value of one. The Pareto front is therefore a
line connecting the points $(0, 1)$ and $(1, 0)$. We use a reference
point (cut off) of $(10, 10)$. This means that all hypervolume values
are between $0$ and $100 - \frac12 = 99.5$.

\subsection{Performance of (1+1)-LM-MA-ES}

For answering the first question we ran experiments on essentially the
same setup as in \cite{loshchilov2018large}. This is mostly a sanity
check since good performance on single-objective problems is only a weak
indicator of the ability to solve multi-objective problems. We tested
(1+1)-LM-MA-ES on the problems Sphere, Ellipsoid, Cigar, Discus, Sum of
Different Powers, and on the Rosenbrock function. Their definitions are
given in table~\ref{table:objectives-so}. The first four problems are
quadratic functions, however, only the Sphere and Cigar functions are a
perfect match for a limited memory algorithm, while Ellipsoid and Discus
profit from adapting a full-rank covariance matrix. The Sum of Different
Powers problem can be seen as an extreme variant thereof where the
conditioning number grows indefinitely as optimization progresses.
Finally, the well-known Rosenbrock function poses an interesting
combination of challenges.

\begin{table}
\begin{center}
	\begin{tabular}{cc}
		\hline
		\textbf{Function} & \textbf{Definition} \\
		\hline
		Sphere & $f(x) = \sum_{k=1}^n x_k^2$ \\
		Ellipsoid & $f(x) = \sum_{k=1}^n 10^{6 (k-1)/(n-1)} x_k^2$ \\
		Cigar & $f(x) = x_1^2 + \sum_{k=2}^n 10^6 x_k^2$ \\
		Discus & $f(x) = 10^6 x_1^2 + \sum_{k=2}^n x_k^2$ \\
		Different Powers & $f(x) = \sum_{k=1}^n |x_k|^{2+4(k-1)/(n-1)}$ \\
		Rosenbrock & $f(x) = \sum_{k=1}^{n-1} 100 \cdot (x_{k+1} - x_k^2)^2 + (x_k - 1)^2$ \\
		\hline
	\end{tabular}
	\caption{\label{table:objectives-so}
		Objective functions used in the single-objective experiments.
	}
\end{center}
\end{table}

\begin{figure}
	\begin{center}
	\includegraphics[width=0.33\textwidth]{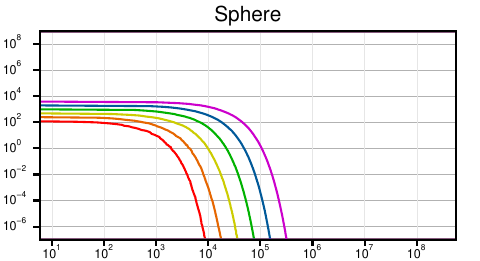}~\includegraphics[width=0.33\textwidth]{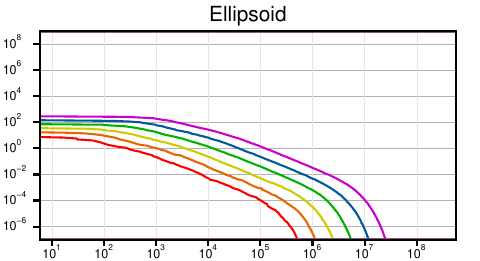}~\includegraphics[width=0.33\textwidth]{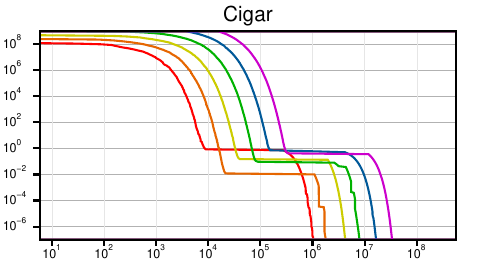}
	\\
	\includegraphics[width=0.33\textwidth]{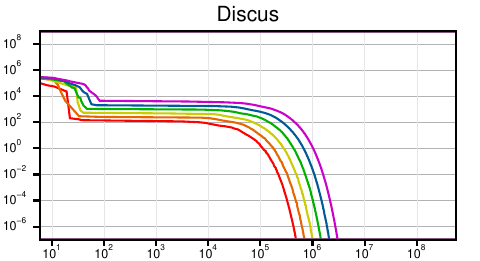}~\includegraphics[width=0.33\textwidth]{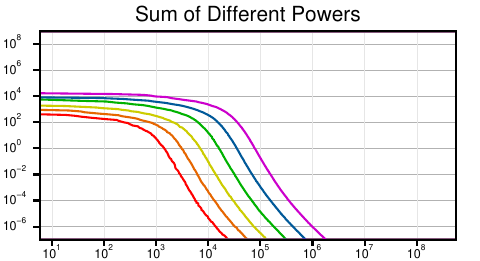}~\includegraphics[width=0.33\textwidth]{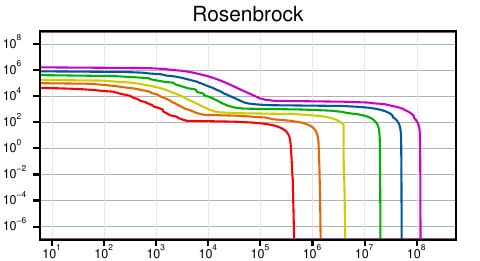}
	\end{center}
	\caption{\label{figure:so}
		Median fitness difference $f(x) - f^*$ over number of function
		evaluations for (1+1)-LM-MA-ES in dimensions 128 (red); 256
		(orange), 512 (yellow), 1024 (green), 2048 (blue), and 4096
		(magenta). Both axes use a logarithmic scale.
	}
\end{figure}

Figure~\ref{figure:so} presents median fitness over five runs of
(1+1)-LM-MA-ES on different problem instances in dimensions
$n \in \{128, 256, 512,$ $1024, 2048, 4096\}$. The Rosenbrock problem is
multimodal, exhibiting a sub-optimal local optimum in high dimensions.
Since we aim to measure convergence speed we dropped runs getting
trapped in the local optimum.

Overall, the performance curves are very similar to the ones reported in
\cite{loshchilov2018large} for the population-based counterpart. In some
cases, (1+1)-LM-MA-ES achieves speedups. The differences may well be
within the margin of what can be achieved by extensive parameter tuning.
Therefore, we consider the differences non-essential. On the quadratic
problems the scaling with problem dimension is linear, which it the best
achievable scaling according to theory \cite{teytaud2006general}.
We conclude that the novel elitist variant is a worthwhile alternative
to the original population-based method. Its use as a building block for
multi-objective optimization is therefore well justified.

\subsection{Performance of MO-LM-MA-ES on Benchmark Problems}

We assess the second and the third question based on test problems with
analytically known Pareto sets and Pareto fronts. A suite of such
bi-objective problems was proposed in \cite{glasmachers2019challenges}.
That benchmark suite consists of 54 quadratic bi-objective problems.
They are designed in such a way that the Pareto set is a line segment
while the corresponding front is either a line segment or a
well-controlled convex or concave curve. The problems differ in whether
the Pareto set is axis-aligned or not, in the conditioning of the
Hessians of the single objectives, and in whether or not the Eigen
vectors of their Hessians are aligned with the coordinate system
(separability) or with each other, or not. Knowing Pareto front and
Pareto set analytically allows to assess a large number of quality
metrics.

In our setup, there is no need to consider all 54 benchmark problems.
First of all, we consider only the 18 problem with linear Pareto fronts
since they allow us to calculate the exact optimal dominated
hypervolume. We further narrow the collection down to 9 problems since
there is no need to distinguish between problems for which the Pareto
set is aligned to a coordinate axis or not, due to rotation invariance
of MO-LM-MA-ES. The relevant properties of the benchmark problems are
listed in table~\ref{table:objectives-mo}.

\begin{table}
\begin{center}
	\begin{tabular}{ccccc}
		\hline
		\textbf{ID} & \textbf{Functions} & \textbf{Aligned} & \textbf{Axes} & \textbf{Hessian} \\
		\hline
		1           & two spheres        & yes              & yes           & same             \\
		2           & sphere + ellipsoid & yes              & yes           & different        \\
		3           & two ellipsoids     & yes              & yes           & same             \\
		4           & two ellipsoids     & yes              & yes           & different        \\
		5           & sphere + ellipsoid & mixed            & no            & different        \\
		6           & two ellipsoids     & mixed            & no            & different        \\
		7           & two ellipsoids     & no               & yes           & same             \\
		8           & two ellipsoids     & no               & yes           & different        \\
		9           & two ellipsoids     & no               & no            & different        \\
		\hline
	\end{tabular}
	\caption{\label{table:objectives-mo}
		Objective functions used in the multi-objective experiments.
		``Aligned'' indicates whether the axes of the ellipsoids are
		aligned with the coordinate system. ``Axes'' indicates whether
		both ellipsoids share the same axes. ``Hessian'' indicates
		whether the Hessians of the two objectives agree (otherwise a
		different covariance matrix is needed at each point of the
		Pareto front).
	}
\end{center}
\end{table}

To the best of our knowledge there does not exist a competing evolution
strategy for solving high-dimensional multi-objective problems. However,
provided enough compute, one can of course run a multi-objective
evolution strategy adapting the full covariance matrix at least on
problems of moderate dimensionality. That is a relevant baseline since
it showcases the different trade-offs in term of algorithm internal
(sampling and parameter updating) and external (fitness evaluation)
complexity. We therefore implemented a full-rank variant of the
algorithm as a baseline method, which is essentially equivalent to
MO-CMA-ES. We made sure that it differs from MO-LM-MA-ES only in
maintaining a full-rank covariance matrix (using an efficient rank-one
update \cite{krause2015more}), as well as the parameter settings
proposed in~\cite{igel2007covariance}. The code overlap between the two
implementations is $\approx90\%$. Therefore, the direct comparison of
measured runtimes is meaningful.

\begin{figure}
	\begin{center}
	\includegraphics[width=0.33\textwidth]{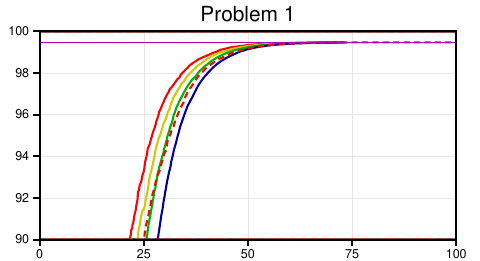}~\includegraphics[width=0.33\textwidth]{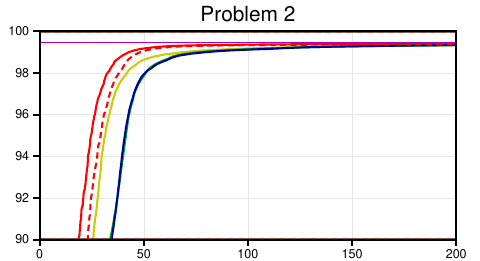}~\includegraphics[width=0.33\textwidth]{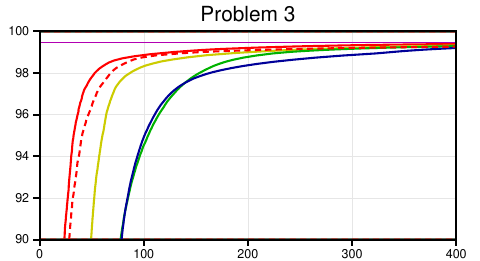}
	\\
	\includegraphics[width=0.33\textwidth]{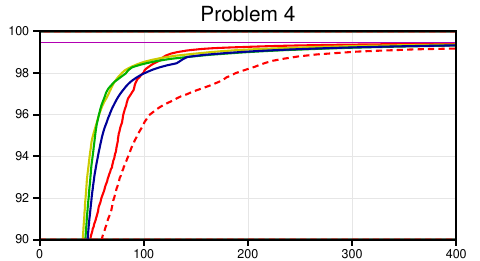}~\includegraphics[width=0.33\textwidth]{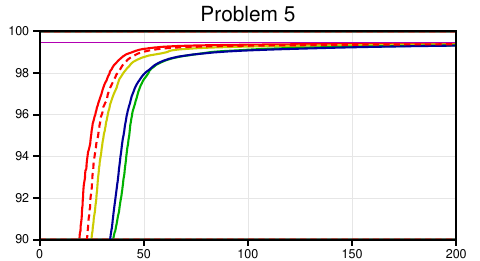}~\includegraphics[width=0.33\textwidth]{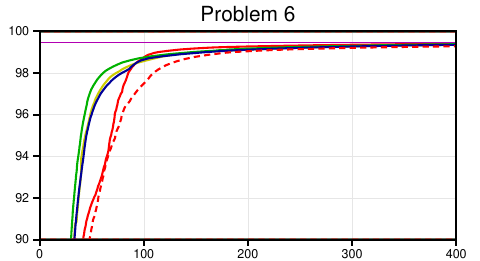}
	\\
	\includegraphics[width=0.33\textwidth]{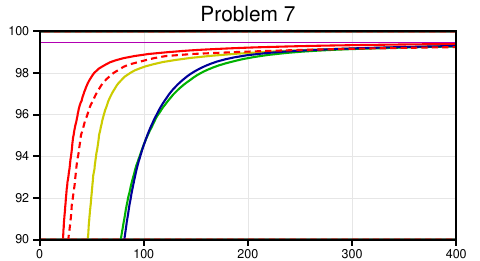}~\includegraphics[width=0.33\textwidth]{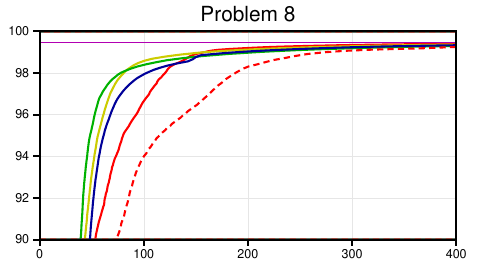}~\includegraphics[width=0.33\textwidth]{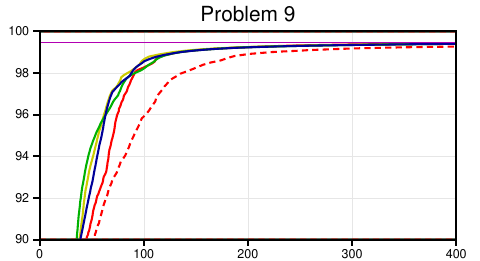}
	\end{center}
	\caption{\label{figure:mo}
		Dominated hypervolume of MO-LM-MA-ES on all nine problems in
		dimensions 128 (red), 256 (yellow), 512 (green) and 1024 (blue).
		The red dashed curve refers to adaptation of the full covariance
		matrix in dimension 128. The horizontal axis is the number of
		function evaluations, normalized (divided) by population size
		and problem dimension.
	}
\end{figure}

Figure~\ref{figure:mo} shows the dominated hypervolume over the number
of fitness evaluations for all nine problems in dimensions 128, 256,
512, and 1024, as well as performance of the baseline in dimension~128.
We observe that all problems are solved to high precision, in all
problem dimensions. It turns out that MO-LM-MA-ES systematically
outperforms the full-rank algorithm. It is not only faster in terms of
wall-clock time, but also in terms of the number of function
evaluations. This is a surprising result. The only systematic advantage
of the low-rank algorithm is that it can afford larger learning rates in
its covariance matrix update due to the reduced number of parameters. We
therefore attribute the performance difference to this property.

It is clear from conceptual considerations that the algorithm internal
complexity of $\Theta(n \log(n))$ per sample for MO-LM-MA-ES scales
favorable compared to $\Theta(n^2)$ per sample for a full-rank
algorithm. We measured the practical effect of this difference by
running both algorithm for a small budget of $10 \cdot \mu \cdot n$
function evaluations on the 512-dimensional Spheres problem with a
population size of $\mu=20$. The experiments were conducted on a single
core of an otherwise idle compute server equipped with Intel Xeon Silver
4314 CPUs running at 2.4GHz. The experiment took 26.85 seconds for the
low-rank algorithm and 76.36 seconds for the full-rank algorithm. This
is a speed-up of nearly factor three in wall-clock time. Although many
steps (like fitness evaluation, non-dominated sorting and
hypervolume-based ranking) are the same for both methods, the runtime
difference clearly demonstrates the computational benefit of the
low-rank approach.

\subsection{Pareto Front Approximation Quality}

The optimal $\mu$-distribution is the optimal placement of a population
of $\mu$ points maximizing the dominated hypervolume for a given problem
\cite{auger2009theory}. For the ``linear front'' variants of the
problems from \cite{glasmachers2019challenges} the optimal
$\mu$-distribution is well known. Provided that the reference point is
sufficiently far from the front the end points of the line segment
forming the Pareto set are attained. The remaining $\mu - 2$ points are
equally spaced in between. Using our reference point, the optimal
achievable hypervolume is then $100 - \frac12 - \frac{1}{2(\mu-1)}$.
When investigating approximation quality we compare actual populations
during a run with this optimal placement. We refer to the difference in
dominated hypervolume as the \emph{hypervolume gap}.

We put particular emphasis on two selected benchmark problems. The easy
problem (problem 1 in the terminology of
\cite{glasmachers2019challenges}) consists of two (shifted) sphere
functions. The hard problem (problem 9) consists of two ellipsoids for
which the Hessians are neither aligned with the extrinsic coordinate
system nor with each other. We refer to the two problems as
\emph{Spheres} and \emph{Ellipsoids}.
The reason for this choice is that the Spheres problem can be solved
without adapting the covariance matrix at all, while Ellipsoids is a
very hard (yet unimodal) problem that requires extensive covariance
matrix adaptation.

\begin{figure}
	\begin{center}
	\includegraphics[width=0.49\textwidth]{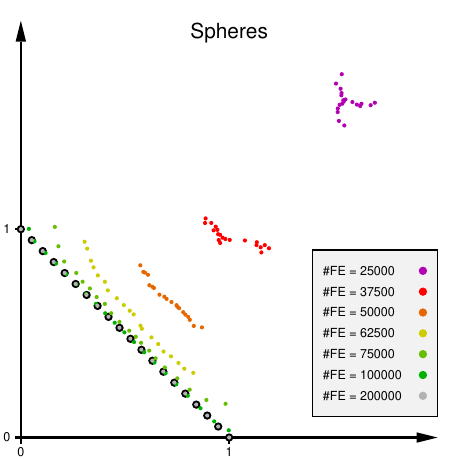}~\includegraphics[width=0.49\textwidth]{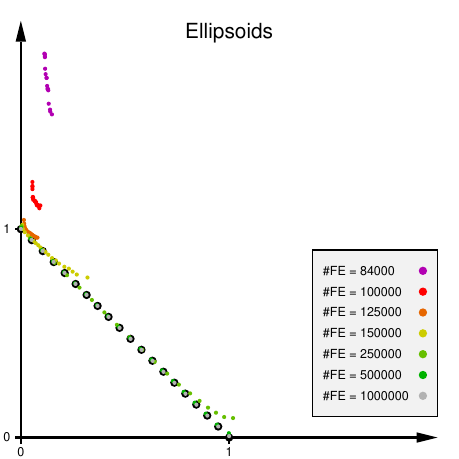}
	\end{center}
	\caption{\label{figure:front-approximation}
		Convergence of the population to the optimal $\mu$-distribution.
		The black dots indicate the optimal $\mu$-distribution for
		$\mu=20$. It can be observed for both problems that the
		population (in objective space) converges to the optimal
		configuration.
	}
\end{figure}

We let MO-LM-MA-ES run with a budget of $1000 \cdot \mu \cdot n$
objective function evaluations on Spheres and Ellipsoids. The runs where
stopped at a target hypervolume gap of $10^{-8}$, which corresponds to
an extremely precise solution.
Figure~\ref{figure:front-approximation} shows the evolution of the
resulting Pareto fronts for $n=128$ dimensions and $\mu=20$ points. It
displays fronts at selected generations. Naturally, early populations
are far from the front. The approach paths in objective space differ
quite significantly between the two problems. In late generations, the
population covers the optimal $\mu$-distribution quite precisely,
although for Spheres less than 10\% of the budget is spent, while the
final state displayed for Ellipsoids corresponds to $\approx 40\%$ of
the budget.

\begin{figure}
	\begin{center}
	\includegraphics[width=0.5\textwidth]{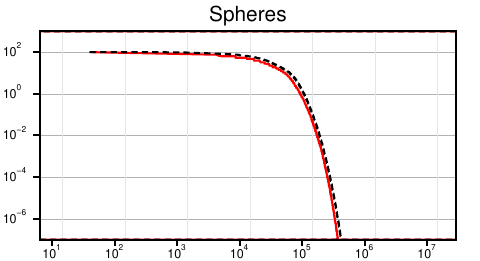}~\includegraphics[width=0.5\textwidth]{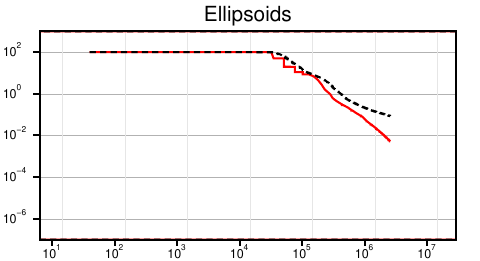}
	\end{center}
	\caption{\label{figure:mo-convergence}
		Log-log-plot of the hypervolume gap over the number of function
		evaluations. The red and solid curve corresponds to the low-rank
		algorithm, while the black dashed curve corresponds to adapting
		a full-rank covariance matrix.
	}
\end{figure}

Figure~\ref{figure:mo-convergence} shows the evolution of the
hypervolume gap. We note that a hypervolume gap of $10^{-2}$ corresponds
to an essentially precise solution as seen in the late states of
figure~\ref{figure:front-approximation}. The logarithmic scale of the
vertical axis resolves even much smaller deviations, which are achieved
with reasonable budgets for the Spheres problem.

Once more, the low-rank algorithm outperforms adaptation of the full
covariance matrix on both problems. While performance of the full-rank
and low-rank algorithms is nearly the same on the Spheres problem, the
low-rank approach is superior on the Ellipsoids problem.

\subsection{Discussion}

We are now in the position to answer our research questions.

\begin{enumerate}
\item
	We find that the new elitist LM-MA-ES is a viable alternative to the
	population-based algorithm.
\item
	Our main question concerns the performance of MO-LM-MA-ES on
	high-dimensional problems. We find that using a low-rank covariance
	matrix results in a speed-up over using a full-rank model. While a
	speed-up in wall clock time is expected due to the reduced algorithm
	internal complexity, to our surprise we also found a speed-up in
	terms of the number of function evaluations. Both effects multiply,
	yielding an overall superior algorithm.
\item
	In all of our experiments MO-LM-MA-ES approximates the optimal
	$\mu$-distribution to high precision. Again, it gets there faster
	than the full-rank algorithm.
\end{enumerate}

Convergence to the optimum is a highly desirable property of every
optimization method in continuous space, yet convergence is rarely
studied for MOEAs. We find that the MO-LMMAES converges to the optimal
$\mu$-distribution. In other words, it approximates the optimal
$\mu$-distribution with arbitrary good quality, of course, provided a
sufficiently large evaluation budget.
We cannot directly visualize the corresponding Pareto sets in
high-dimensional search spaces. However, we can conclude from the fact
that Sphere as well as Ellipsoid are convex quadratic functions encoding
squared (Euclidean and non-Euclidean) norms that convergence in
objective space implies converge of the population to the optimal
$\mu$-distribution in search space.

It is unclear whether the effect that the low-rank model outperforms the
full-rank model carries over to the limit of very large compute budgets.
A principled argument against this is that the full-rank model allows
for a better convergence rate. That's an asymptotic effect, but it
should kick in eventually. However, it does not show up in our
experiments. This may indicate that multi-objective evolution strategies
hardly ever enter the ``late'' convergence phase, in particular for
high-dimensional problems. They may rather solve the problem to
sufficiently high precision while still being in the transient phase in
which they keep adapting the covariance matrix. An alternative
explanation is that the low-rank model is indeed sufficient, even if the
inverse Hessians of both objective functions are incompatible with the
low-rank model. This may happen if the crucial operation in the late
phase of optimization is to move \emph{along} the Pareto set, which is
what a low-rank model is very well suited for. Clarifying this point
requires an in-depth investigation of the optimization behavior. We
leave this task for future research.

\section{Conclusion}

Our main focus was the construction of a class of efficient
multi-objective evolution strategies for high-dimensional problems. They
leverage the generic blueprint of MOEAs with indicator-based selection
by incorporating variation operators suitable for high-dimensional
problems.

We have presented two novel algorithms. As an intermediate step we
developed the (1+1)-LM-MA-ES, an elitist variant of an efficient ES for
high-dimensional problems. It is then used as a building block in a
particular instance of the above construction, namely the MO-LM-MA-ES.

According to our evaluation, the (1+1)-LM-MA-ES works about as well as
its population-based counterpart. It is therefore a suitable building
block for our main algorithm, the MO-LM-MA-ES. That algorithm performs
well on high-dimensional instances from a suite of quadratic
bi-objective problems which critically require covariance matrix
adaptation. The new algorithm successfully solves high-dimensional
ill-conditioned problems to high precision, yielding well distributed
Pareto front approximations.

At the time of this writing there exist no established benchmark problem
collections for high-dimensional multi-objective optimization. It would
be well conceivable that such problems can be constructed by combining
the BBOB/COCO suites for bi-objective and for high-dimensional
optimization.\footnote{\url{https://github.com/numbbo/coco}}
We would highly appreciate such a construction, which would offer the
opportunity for a more standardized performance assessment.

\bibliographystyle{plain}

\end{document}